# MFL_COVID19: Quantifying Country-based Factors affecting Case Fatality Rate in Early Phase of COVID-19 Epidemic via Regularised Multi-task Feature Learning

Po Yang[1†]   Jun Qi[2]   Xulong Wang[3]   Yun Yang[3]

1. Department of Computer Science, The University of Sheffield, Sheffield, United Kingdom,
   po.yang@sheffield.ac.uk
2. Department of Biomedical Engineering, Oxford University, Oxford, UK
3. School of Software, Yunnan University, Kunming, China,

## ABSTRACT

Recent outbreak of coronavirus disease 2019 (COVID-19) has led a rapid global spread around the world. Many countries have implemented timely intensive suppression to minimize the infections, but resulted in high case fatality rate (CFR) due to critical demand of health resources in the early phase of epidemic. Other country-based factors such as sociocultural issues, ageing population etc., has also influenced practical effectiveness of taking interventions to improve morality in early phase. To better understand the relationship of these factors across different countries with COVID-19 CFR is of primary importance to prepare for potentially second wave of COVID-19 infections. In the paper, we propose a novel regularized multi-task learning (ML) based factor analysis approach for quantifying country-based factors affecting CFR in early phase of COVID-19 epidemic. We formulate the prediction of CFR progression as a ML regression problem with observed CFR and other countries-based factors. In this formulation, all CFR related factors were categorized into 6 sectors with 27 indicators. We proposed a hybrid feature selection method combining filter, wrapper and tree-based models to calibrate initial factors for a preliminary feature interaction. Then we adopted two typical single task model (Ridge and Lasso regression) and one state-of-the-art MTFL method (fused sparse group lasso) in our formulation. The fused sparse group Lasso (FSGL) method allows the simultaneous selection of a common set of country-based factors for multiple time points of COVID-19 epidemic and also enables incorporating temporal smoothness of each factor over the whole early phase period. Finally, we proposed one novel temporal voting feature selection scheme to balance the weight instability of multiple factors in our MTFL model. Our model was evaluated in a COVID-19 dataset containing 12 countries across the world and 17 provinces in China. The results show that we explored each indicator's correlation and calculated its weight under current missing, sparse and insufficient data. we discover that Power distance index (PDI), Diabetes mellitus disease mortality rate, gender rate and SPAR_detect capability are important factors associated to CFR over the whole early phase of COVID-19 in a region or country.

## 1 INTRODUCTION

As of 1st April 2020, , the ongoing global epidemic outbreak of coronavirus disease 2019 (COVID-19) has spread to at least 146 countries and territories on 6 continents, resulted in 896 thousands confirmed case and over 45 thousands deaths [1]. Most countries have took timely intensive suppression that immediate lock-down in cities at epicentre of outbreak like in Italy and China [33-35]. This intervention enables efficiently minimizing the infections in the early phase of COVID-19 epidemic, but potentially resulted in high case fatality rate (CFR) on the cases of critical shortage of health resources. Also, owning to huge difference of other country-based factors like culture issues, ageing population, urban density etc., intensive suppression might consequence a widely varied CFR across countries in the early phase of COVID-19 outbreak, such as over 10% in Spain and Italy [2], but less than 4% in Germany and Japan [2]. Thus, it reveals that practical effectiveness of COVID-19 intervention strategies to certain country has been varied in light of many country-based factors including aging population, human mobility, health resources, culture issues, etc. It is crucial but hard to know how these factors affect CFR in the early phase of COVID-19 epidemic tailored to the specific situation in each country.

In order to explore the relationship of these factor with CFR, feature selection [3] is a typical process to automatic or manual selection of these significant features (factors) which contribute most to certain prediction variable (CFR). In these process, the most common techniques include: 1) filter methods that score each feature according to their intrinsic relevance or correlation, set a threshold or the number of thresholds to be selected, like ReliefF [4-5], correlation-based feature selection [6], fast correlated based filter [7], etc.; 2) wrapper methods that measure the usefulness of features based on the classifier performance, such as recursive feature elimination [8], sequential feature selection [9] or genetic algorithms [10]. 3) Embedded and hybrid methods that performs feature selection during the execution of optimising their objective function or performance of a learning model or algorithm, including L1 (Lasso) regularization [13-14] and decision tree algorithms (CART, C4.5, random forest) [11-12] [15]. In the last decades, above feature selection approaches have been utilised in many practical applications with outstanding performance in removing noise features and good versatility. However, one key problem of above approaches is that they require sufficient and well-labelled data or features to ensure the accurate learning task. In the COVID-19 scenario, the CFR and its related country-based factors in the early phase may contain huge uncertainty, sparse and missing data that hardly train a fine model to select features.

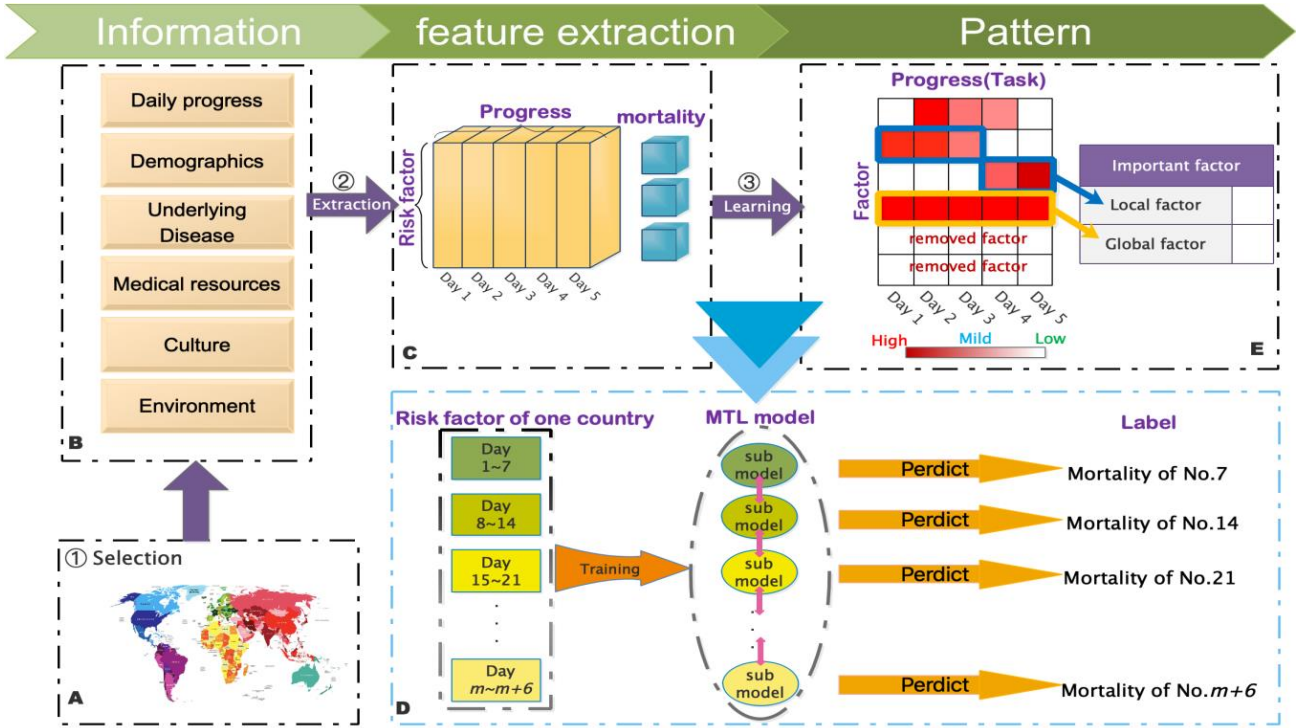

**Figure 1: Schematic illustration of multi-task feature learning for quantifying COVID-19 country-based factors**

Differing with traditional feature selection methods, multi-task feature learning (MTFL) [16-18] approaches have shown a strong capability to handle the problem of data insufficiency. This is the case, when only a few data per task are available, learning or training multiple related tasks simultaneously with shared features can often greatly improve performance relative to learning each task independently. Recently, many existing methods impose the shared features by a generalized $L_1$-norm regularization [19], or more specifically, a joint regularization on the $L_{p,1}$-norm of the learning weights [20], where $p$ can be set to 1, 2, or other values. These methods select the subspace while seeking the weights of the decision functions by minimizing a convex optimisation problem on the sum of the joint regularization and the loss [21-22]. These regularized multi-task learning approaches [23-28] have been used in many Alzheimer's disease progression and deliver promising performance in biomarker feature selections. MTFL techniques has some potential to explore the impacts of factors on COVID-19 CFR.

In the paper, we propose a novel regularized multi-task feature learning (ML) approach for measuring country-based factors affecting CFR in early phase of COVID-19 epidemic. As shown in Fig.1, we formulate the prediction of CFR progression as a ML regression problem with observed CFR and other countries-based factors. In this formulation, all CFR related country-based factors were defined into 6 sectors including COVID-19 progression, demographic information, morbidity of underlying diseases, healthcare resources, international health regulation (IHR) core capability and social-culture with 27 indicators. The prediction of COVID-19 CFR at each week is a task, where multiple prediction tasks at different time points are performed simultaneously to capture the intrinsic temporal relatedness among different tasks.

We first proposed a hybrid feature selection method combining filter, wrapper and tree-based models to calibrate initial factors for a preliminary feature interaction. Then we adopted two typical single task model (Ridge and Lasso regression) [29] and one state-of-the-art MTFL method (fused sparse group lasso) [28] in our formulation. The fused sparse group Lasso (FSGL) method allows the simultaneous selection of a common set of country-based factors for multiple time points of COVID-19 epidemic and also enables incorporating temporal smoothness of each factor over the whole early phase period. Finally, we proposed one novel temporal voting feature selection scheme to balance the weight instability of multiple factors in our MTFL model. Our model was evaluated in a COVID-19 dataset containing 12 countries across the world and 17 provinces in China. The results show that we explored each indicator's correlation and calculated its weight under current missing, sparse and insufficient data. Our key contributions are:

- One regularized MTFL approach is proposed for effectively measuring country-based factors affecting CFR in early phase of COVID-19 epidemic. In comparing to traditional single-task feature selection method [29], it can deal with insufficient and uncertain COVID-19 data, and provide both global and local temporal analytic of these factors over early phase of COVID-19 epidemic.
- One novel temporal voting feature selection scheme is proposed to balance the weight instability of multiple factors in our regularized MTFL approach. We provide an evidence verification point that this scheme can significantly improve the stability of MTFL model in processing complex COVID-19 data.
- By leveraging verification in experimental progress, we also discover and proof that Power distance index (PDI), Diabetes mellitus disease mortality rate, gender rate and SPAR_detect capability are important factors associated to CFR over the whole early phase of COVID-19 in a region or country. It reveals that the culture difference might significantly affect effectiveness of COVID-19 intervention in early phase.

The remainder of this paper is arranged as follows. Section 2 introduces our methodology of building the MTFL model for COVID-19 CRF study. In the Section 3, the materials and implementation of experiment are reported. Section 4 provides detailed experimental evaluation and discussion. The conclusion and future directions are given in Section 5.

## 2 Conceptualisation and data preprocessing

### 2.1 Conceptualisation

In epidemiology, the concept of case fatality rate (CFR) usually refers to the proportion of deaths from a certain disease compared to the total number of people diagnosed with the disease for a certain period of time. A CFR is conventionally expressed as a percentage and represents a measure of disease severity. As for COVID-19, due to high transmission rate and insufficiency of test kits in early phase, it is hard to identify real infectious population. Thus, here, we would use a common definition of daily CFR with confirmed death over confirmed infections every day, as shown in Equation (1).

$$CFR = \frac{Confirmed\_Death}{Confirmed\_Infection} \quad (1)$$

From this equation, it shows that the CFR has a direct relationship with COVID-daily progression, which was defined as the first sector as shown in the Table.1

In recent studies [1] [29-30], it revealed that timely suppression in the early phase of COVID-19 epidemic can minimize the infections in many countries, but result in high (CFR). In order to study CFR associated factors, we searched PubMed with the search terms 'covid-19', 'death cause', 'factor importance', 'death prediction', and found out other CFR associated factors: 1) The demographics includes the rate of aging population, gender and smoking. 2) The morality of key diseases in a country refer to several identified diseases associated to COVID-19 CFR, including lung cancer, coronary heart disease, diabetes, hypertension. 3) The healthcare resources includes rate of the number of hospital beds, the number of physicals per million people, and health quality index. As for social and behavioural issues of a region affecting the CFR, WHO has measured each country to prevent, detect, and respond to events are essential to control public health risks called International Health Regulations (IHR), where it was categorized in our factors as well. We have also considered to quantify social-culture index using Hofstede's cultural dimensions theory [32], including power distance index, individualism, masculinity, etc. Finally, air quality index was also taken into account in our model.

**Table 1. Country-based factors affecting COVID-19 CFR**

| | |
|---|---|
| **COVID-19 progression** | Daily new cases |
| | Daily new deaths |
| | Current active case |
| | Cases per million inhabitants |
| **Demographics** | Aging population rate (over 65) |
| | Gender rate |
| | Smoking rate |
| **Mortality of key diseases** | Lung cancer mortality rate |
| | Coronary heart disease mortality rate |
| | Diabetes mellitus disease mortality rate |
| | Hypertension disease mortality rate |
| **Healthcare resource** | Number of hospital beds per million people |
| | Number of confirmed diagnoses per million people divided by beds rate |
| | Number of physicians per million people |
| | Healthcare access and quality index |
| **IHR (International health regulation) core capacity index** | Capacity to prevent the IHR state party annual reporting |
| | Capacity to detect the IHR state party annual reporting |
| | Capacity to respond the IHR state party annual reporting |
| | Capacity to enabling function of the IHR state party annual reporting |
| | Capacity to operational readiness of the IHR state party annual reporting |
| **Social-culture index** | power distance index |
| | individualism |
| | masculinity |
| | uncertainty avoidance index |
| | Number of arrivals to the country / province each year (liquidity) |
| **Others** | Air quality index |

### 2.2 Data collection and preprocessing

Considering stability and robustness of the prediction model, we chose the regions with real reported death toll at 42 consecutive days from the first confirmed COVID-19 case, totally 12 countries (France, Germany, Italy, Japan, Singapore, South Korea, Spain, Sweden, United Kingdom, China and Philippines) across the world and 17 provinces in China. 6 factors of interest covering total 27 potential indicators of progression are selected that may affect the CFR of the COVID-19 outbreak. Notably here, the reason that we only use the first 42 days of COVID-19 epidemic is that many countries have implemented suppression during the 35-45$^{th}$ day, as a consequence of significant changes of our defined factors, like health resources. After taking suppression for 2-3 weeks, social-culture issues have limited impacts on social distance measures in most of countries.

We collected factors associated data sources mainly from worldmeter, WHO, world-bank and internal database from Oxford University. There are various problems with the raw data collected, such as missing values, outliers, etc. Most data sources do not have structured data. In order to allow our MTFL model to work properly, we have done some data preprocessing with following steps: 1). to clarify and unify the attributes of data from multiple resources for matching associated factors. 2) to remove or merge data with repeated attributes. 3) To identify and remove some apparent outliners. 4) To smooth and remove some noise data. Notably, in the process of collecting data, it is inevitable that some sampling data were missed. The average method was used to fill in the data in order to minimize its impact on MTFL learning.

## 2.3 Preliminary feature processing

When converting initial CFR factor associated data into reliable and accurate features, we proposed a hybrid feature selection method combining filter, wrapper and tree-based models to calibrate initial factors for a preliminary feature interaction. Firstly, we made a standard scaling process to all data, and use simple univariate features to make a preliminary selection. Then, we applied Pearson Correlation and F-score to measure the linear correlation between each feature and the CFR. Based on these two filters, we could effectively remove some noise features and keep good versatility. But only filter methods cannot guarantee that the selected feature subset is the optimal feature subset that matches the subsequent machine learning algorithm. Then, we have applied three wrapper methods (recursive feature elimination (RFE), random forest and XGBoost) to train the features. After each round of training, the features corresponding to several weight coefficients are eliminated, and then the next round of training is performed based on the new feature set. We use RFE to exhaustively delete a feature that has the least negative impact on the model in the existing feature subset until the number of selected features meets the requirements. Random Forest and XGBoost are tree-based machine learning models. These non-parametric tree models record how each variable gradually reduces the model loss in the bifurcation of tree nodes during the establishment process and can analyze the feature importance of each feature based on the above records, we can delete some unimportant variables based on the importance ranking. After taking above preliminary feature processing, the final training data set is to take the intersection of features selected by the first two coefficients, plus the intersection of features selected by the three models. The combination of multiple methods enhances robustness. It is worth noting that all the above feature selection schemes are operated under the setting of a single task, that is, every time point.

## 3 Multi-task feature learning

### 3.1 Problem formulation

In order to simultaneously train above a variety of CFR associated factors, a multi-task feature learning model was constructed, as shown in Fig.1. Consider a multi-task learning of $k$ tasks with $n$ training samples of d features. Let $\{x_1, x_2, \ldots, x_n\}$ be the input data for the samples, and $\{y_1, y_2, \ldots, y_n\}$ be the predicted value for each sample, where each $x_i \in \mathbb{R}^d$ represents the feature data of a CFR associated factor, and $y_i \in \mathbb{R}^k$ is the predicted value of CFR in a certain region or country. The formula for a linear regression model is given by $f_i(x) = x^T w_i$ where $w_i$ is the weight vector of the model. Here in ML, a matrix representation is performed and facilitates intuitive understanding of algorithm and actual programming operations.

Then, let $X = [x_1, \cdots, x_n]^T \in \mathbb{R}^{n \times d}$ be the data matrix, $Y = [y_1, \cdots, y_n]^T \in \mathbb{R}^{n \times k}$ be the predicted matrix, and $W = [w_1, w_2, \ldots, w_k] \in \mathbb{R}^{d \times k}$ be the weight matrix. The process of establishing a MTL model is to estimate the value of $W$, which is the parameter to be estimated from the training samples. One common paradigm for regression in MTL is to minimize the following objective function as below:

$$\min_W L(Y, X, W) + \lambda R(W) \quad (2)$$

where $L(Y, X, W)$ means the empirical loss on the training set, $R(W)$ is the regularization term that encodes task relatedness and $\lambda$ is the regularization parameter.

In order to solve above optimisation problem, many prior work that model relationships among tasks using regularizations methods. Normally, they assume the loss to be square loss and common regularization terms are L1 norm and L2 norm, which are separately named as *Lasso regression* [22] and *ridge regression* models [22]. These two models are well-known sparse representation linear algorithms, which simultaneously performs feature selection and regression. One limitation of the regression model above is that the tasks at different time points are supposed to be independent with each other, which is not the case in the temporal and multifactorial study considered in this article.

### 3.2 Temporal supporting group structure

As shown in Equation 1, Daily CFR is as an indicator directly associated with daily confirmed death and cases of COVID-19. But these two values are accumulated to the day. Considering that in our MTFL model, the prediction of CFR at a single time point as a regression task, we formulate the progression of crucial mortality as a multi-task regression problem. Here, when multiple tasks are cascaded together, the missions of some tasks are unified, each task is no longer limited, and they work together to predict critical time points. Therefore, the selection and definition of critical CFR time point are crucial to the performance of our MTFL model.

If the daily CFR is employed as the prediction for daily tasks, the model will lose stability. One nearly universal truth is that an epidemic disease stops spreading, and the mortality rate will be a fixed value. However, during the propagation process, this value is constantly shaking, but it will eventually converge. This phenomenon is similar to the stochastic gradient descent process in optimization theory. Thus, for improving the generality of our MTFL model, we consider a group structure which enables coupling the related tasks using one common task target of daily CFR (critical time point). More precisely, we assume that the tasks are divided into groups, and each group has 7 tasks. In the group, the learning objective of the seventh task is the joint learning target of all tasks. Between groups: the key mortality time point shared within group support the continuous progress between the groups. The framework can be flexibly applied to the real-world scene of the epidemic, as shown in Fig.1.D.

This partially single-output multi-task structure is similar to the optimization of mini-batch gradient descent (MBGD). In the optimization, MBGD takes into account the advantages of batch gradient descent and stochastic gradient descent, which avoids the inefficient strategy of using overall samples and the oscillation caused by the use of a single sample. Analogously, the structure we proposed in this paper not only avoids the use of a single target for all tasks, resulting in the same learned pattern for all tasks, but also reduces the model oscillation caused by the use of own target for each task, thereby improving the robustness of the model.

### 3.3 Structural regularization method

Structural regularization methods in MTL constrains optimization by using regularization terms and shares information between tasks. In the research of Alzheimer's disease (AD) progress, MTL has made outstanding contributions, there are many prior work [21-29] that model relationships among tasks using novel regularization. In order to solve our regression model, we have main considered two typical single task model (Ridge and Lasso regression) [29] and one state-of-the-art MTFL method (fused sparse group lasso) [28] in our formulation.

*Ridge regression* constrains variables to a smaller range for reducing some factors with little impacts on model's prediction. Unfortunately, this reduction means that these variables are still considered. To solve this problem, *Lasso* was proposed as a new sparse representation linear algorithm, which simultaneously performs feature selection and regression. Some variables are set to zero directly to achieve sparsity and dimensionality reduction.

$$\min_{W} L(Y, X, W) + \lambda \| W \|_1 \quad (2)$$

$$\min_{W} L(Y, X, W) + \lambda \| W \|_2 \quad (3)$$

The fused sparse group Lasso (FSGL) method [28] allows the simultaneous selection of a common set of country-based factors for multiple time points of COVID-19 epidemic and also enables incorporating temporal smoothness of each factor over the whole early phase period. FSGL involves sparsity between tasks, where it considers both common features at different points in time and unique features to each task. This feature is helpful to improve the overall performance of the model. cFSGL formulation solves the following convex optimization problem:

$$\min_{W} \| Y - XW \|_F^2 + \lambda_1 \| W \|_1 + \lambda_2 \| RW^T \|_1 + \lambda_3 \| W \|_{2,1} \quad (5)$$

where the first term measures the empirical error on the training data, $\|W\|_1$ is the lasso penalty, $\|RW^T\|_1$ is the fused lasso penalty, and $\|W\|_{2,1}$ is the group lasso penalty.

Lasso and group lasso combined employ is called sparse group lasso, which allows simultaneous selection of a common feature for all time points and internally generates sparse solutions in response to different time points. Fused lasso penalty having a given temporal smoothness, which makes selected features at nearby time points similar to each other. In addition, notice that FSGL's formula involves three non-smooth terms. We propose to solve this optimization problem using accelerated gradient descent method. As shown in Fig.3, we demonstrated the key difference of single task models (Lasso regression) [29] and MTL model (FSGL) [28] as targeted methods.

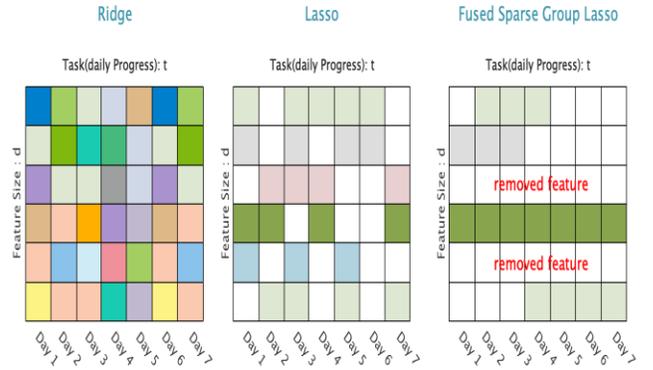

**Fig.3. Difference between Lasso [29] and FSGL [28]**

Notably, in Fig.3, it appears that each column represents the weight of one task model, but the difference is that the vision of Lasso only covers the current model itself, and the learning of each weight of FSGL takes into account the relationship with other tasks. From the distribution of weights in matrix, FSGL demonstrates a meaningful and regular temporal pattern along 7 days than Lasso. FSGL's constraint rules are executed simultaneously on the weights of multiple tasks at the same time. First, a feature can be captured when all tasks are important or not, e.g. Lines 4 and 5. This is a common factor in all tasks. In addition, the specificity of each task model can also be captured, e.g. except for the first feature, most weights of Day1 and Day2 are the same. The structure of FSGL combined with temporal smooth constraints has achieved excellent effects in the study of Alzheimer's disease progression, and this property is once again applied to our framework to fit the tendency of the epidemic. The pattern fitted by the model mines a subset of features that play a role as the epidemic development.

### 3.4 Temporal voting feature selection

While FGCS based MTFL algorithms help analyse the importance of CFR associated factors, one important issue in our experiments is that due to huge uncertainty of COVID-19 dataset, the above MTFL algorithm perform instable. This phenomenon may be related to the randomness of the algorithm itself to the data. In order to unravel this problem, we scheme to repeat the experiment multiple times. Under the guidance of the law of large numbers, the stability of features can be captured. Thus, we proposed one novel temporal voting feature selection scheme to balance the weight instability of multiple factors in our MTFL model.

As shown in Fig.4, the feature sets are submitted to the voting center to be summarized. The voting center is a global feature count table that maintains to record the number of times a feature has been selected by all tasks. The cumulative counter polls among tasks to perform counting operations. The statistical results sort the global feature count table and take the features of the top numbers as the result of feature selection. Specifically, for task $i$ feature: subset $j$, we detect each feature $i$:

$$s_i = \begin{cases} 1, feature_i\_in\_subset_i \\ 0, feature_i\_not\_in\_subset_i \end{cases} \quad (6)$$

For global cumulative counter $C$:

$$c_i = c_i + s_i \quad (7)$$

Practically, our voting mechanism will be employed twice in the MTFL. Firstly, in an experiment, we focused on the weight of the trained model, and counted the importance of the weight (voting), and then ranked according to the importance (ranking). Secondly, we aim at the ranking obtained from all the experiments in the previous step, and sum up to obtain the total ranking importance (voting), and then rank according to the total importance (ranking).

Once the stability feature ranking is obtained, it can be used to reversely find the best experimental model. It is worth noting that relaxing conditions are allowed, for example, part of important features (top features) in the best experiment match the ranking of stable features.

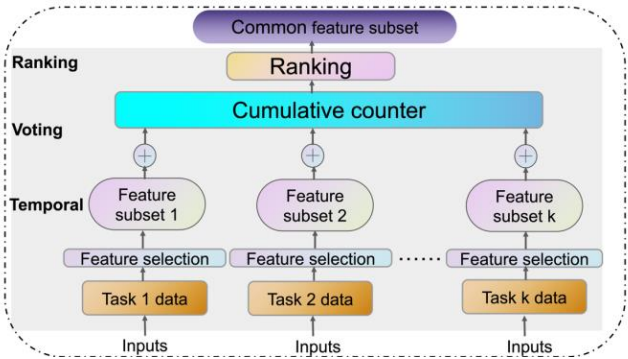

**Fig.4. Difference between Lasso [29] and FSGL [28]**

## 4 EXPERIMENTS

### 4.1 Parameter settings

The goal of multi-task learning is to improve the learning performance of the algorithm by making use of the correlation between different tasks. In the following, the model performance based on this assumption is verified. During each training session, dataset was randomly split into training and testing sets using a ratio 9:1, i.e., models were built on 90% of the data and evaluated on the remaining 10% of the data. Models parameters were selected by 5-fold cross validation. The mean and standard deviation in results based on 100 iterations of experiments on different splits of data. The evaluation metric of model is the modified rMSE with multi-task model estimation. The detailed mathematical definition is as follows:

$$rMSE = \frac{\sum_{i=1}^{t} length(i) * \sqrt{\frac{(Y(i)_{\_hat} - Y(i))^2}{length(i)}}}{\sum_{i=1}^{t} length(i)} \quad (8)$$

where the loss for the t task is calculated separately, and finally added together and divided by the total task sample. Length (i) denotes the number of samples in the i-th task, Y (i) represents the actual value of the sample, and Y (i) _hat denotes the predicted value of the sample.

We employed sample data from 32 countries and regional data from more than 20 provinces in mainland China. All countries / regions were observed for 42 days at the beginning of the outbreak, corresponding to 42 sub-task models. As mentioned above, every 7 tasks are a learning stage. In terms of the progress of the entire project, 42 days can be divided into 3 sub-phases to observe the phenomenon: prior period, middle period and last period. Multi-task learning method was reproduced and compared with two single-task learning methods (ridge regression and Lasso regression) at the same time.

### 4.2 Global importance of CFR factors

Our first experiment is to evaluate the global importance of CFR factors over 42 days in the early phase of COVID-19 using single task model (Ridge, Lasso) and MTFL (FSGL) model. After running 100 times of repeated experiments, Table.2 shows the performance of model accuracy and stability. It shows that our proposed MTFL is more stable than single task regression model, where its feature analysis is more convinced in a global setting.

**Table2: Model performance comparison**

| Experiment | Model | rMSE |
|---|---|---|
| (a) | Ridge | 150.08±11.34 |
| (b) | Lasso | 146.98±13.30 |
| (c) | FSGL | 91.95 ± 6.69 |

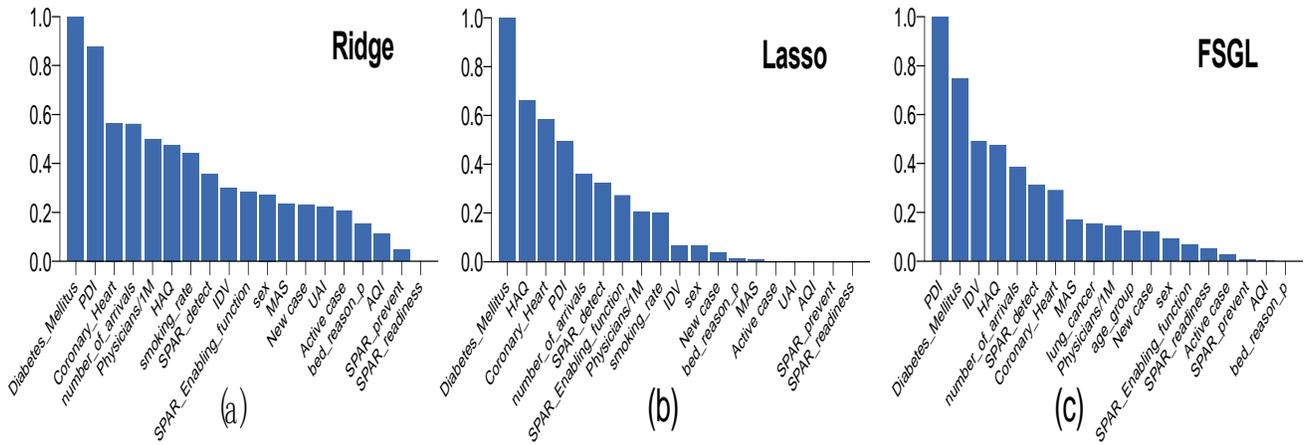

**Fig.5. Global importance of CFR factors using Ridge, Lasso and FSGL**

Fig.5 demonstrates the importance of CFR associated factors over the whole 42 days. These three training models show similar results that the most important factors associated to CFR in a region or country are Power distance index (PDI), Diabetes mellitus disease mortality rate, gender rate and SPAR_detect capability.

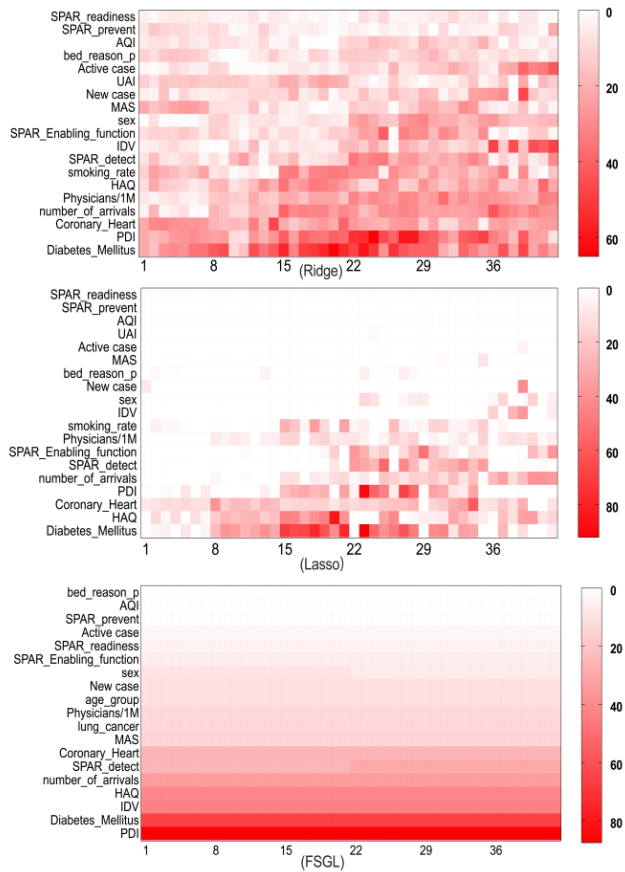

**Fig.5. Local importance of CFR factors using Ridge, Lasso and FSGL**

### 4.3 Local importance of CFR factors

Our second experiment is to evaluate the local importance of CFR factors over 42 days in the early phase of COVID-19 using single task model (Ridge, Lasso) and MTFL (FSGL) model. Similarly, we use the model training results from Table.2 to generate the heat-map of three models in Fig.5. The results show that single task model (Ridge and Lasso) cannot demonstrate meaning or regular temporal pattern of CFR associated factors. It means that these models only work on evaluating global importance of CFR factors, but not work on local importance of CFR factors. But our proposed MTFL shows advantages on temporal smoothness, where its feature analysis is more convinced in showing local importance.

### 4.4 Feature ranking stability comparison

As we mentioned, the COVID-19 data in early phases contain huge uncertainty and insufficiency, as a result of MTFL model training instability. In order to better understand the effectiveness of our proposed MTFL model, we have used SEIR model to generate some daily progress COVID-19 data, and taken different experimental protocols to evaluate them: 1) to use only 48 real sample data for training each task. 2) to use 48 real sample data and 7 simulated data for training each task. Above experiments have been run over 100 times. The results are shown in Fig. 6.

The results show that the most important factors associated to CFR over the whole early phase of COVID-19 in a region or country are Power distance index (PDI) and Diabetes mellitus disease mortality rate. It reveals that the culture difference might significantly affect effectiveness of COVID-19 intervention in early phase. In some Asian countries with higher PDI, timely suppression is much easier accepted than these countries with lower PDI. This issue needed to be noticed in preparation of the possible second wave of COVID-19.

Meanwhile, in order to evaluate stability of each sector of CFR associated factor under different setting, we have evaluated them separately, as shown in Table.3. The results show that if we took all CFR associated factors in training our MTFL model (No 1 and 2), the model shows the best accuracy and stability. But using all real data in No 1 gives the best training performance. If we reduce the daily progression factors (No 5), the training accuracy significantly reduced, where it means that the CFR is directly associated with COVID-19 daily progression factors. If we reduce healthcare or social-culture factors, the training accuracy reduced some but it is probably due to reduced data samples.

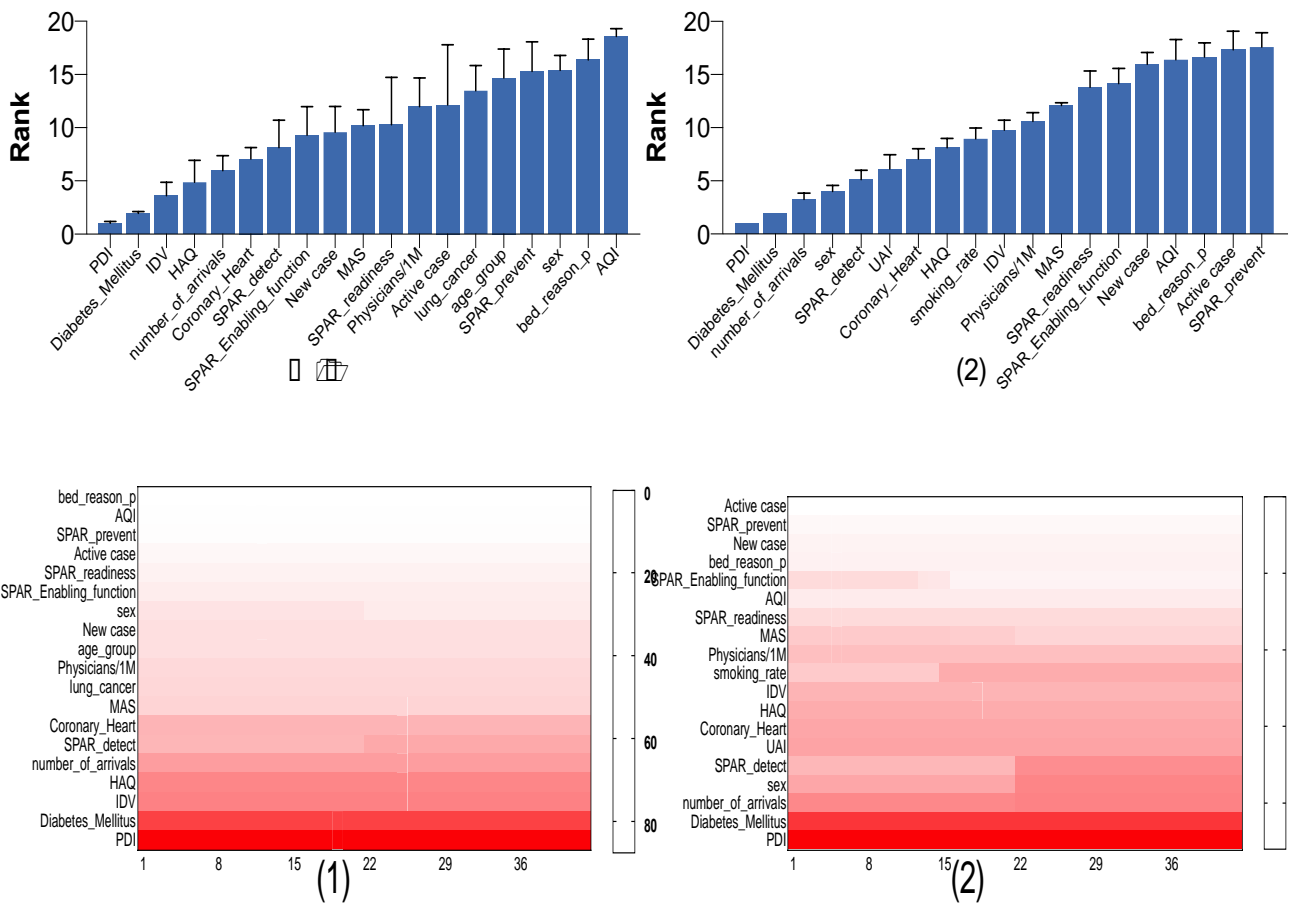

**Fig.6. Global and Local importance of CFR factors using real and simulated data.**

**Table3: Performance comparison using different factors**

| No | Daily Progression | Healthcare | Socil-culture | rMSE |
|----|---|---|---|---|
| 1 | ✔ | ✔ | ✔ | 91.95±6.69 |
| 2 | ✔ | ✔ | ✔ | 99.59±6.57 |
| 3 | ✔ | ✔ |   | 127.12±9.66 |
| 4 | ✔ |   | ✔ | 122.48±9.19 |
| 5 |   | ✔ | ✔ | 145.19±10.48 |

### 4.5. Discussion

There are some limitations to our proposed MTFL model and analysis. First, our model's training depends on the practically collected COVID-19 data with huge uncertainty. The quality of raw dataset greatly impacts on training performance. Due to insufficient sample data, we have to take each province of China as an individual region in the dataset. This might lead to global distribution and balance of dataset, where the data from China shows a strong impact on results. Secondly, it is hard to define an accurate baseline of CRF associated data to train the model. The confirmed infection case of COVID-19 in each country is heavily depended on their screening capability, but not reflects the real infectious population. While we have tried to use SIER model to simulate some confirmed case, the training results show that this simulated data does not help improve stability of training model. We would attempt to improve it in the future work.

# 5 CONCLUSION

In the paper, we propose a novel regularized multi-task feature learning (MTFL) approach for measuring country-based factors affecting CFR in early phase of COVID-19 epidemic. It can deal with insufficient and uncertain COVID-19 data, and provide both global and local temporal analytic of these factors over early phase of COVID-19 epidemic. We believe that this MTFL approach can be used to analysis more factors in future COVID-19 research work.